\documentclass[a4paper, 10pt, conference]{ieeeconf}      % Use this line for a4 paper

\IEEEoverridecommandlockouts                              % This command is only needed if 
\usepackage{graphicx} % for pdf, bitmapped graphics files
\usepackage{color}
\usepackage{algorithm}  
\usepackage{algpseudocode}  
\usepackage{amsmath}
\usepackage{array}
\usepackage{booktabs}
\usepackage{subfigure}
\title{\LARGE \bf
A Novel Autonomous Robotics System for Aquaculture Environment Monitoring$^{\ast}$
}

\author{Tianqi Zhang$^{2\star}$, Tong Shen$^{2\star}$, Kai Yuan$^{2\star}$, Kaiwen Xue$^{2,1}$ and Huihuan Qian$^{1,2\dagger}$ 
\thanks{$^{\ast}$This paper is support by University Stability Support Program from Shenzhen Natural Science Foundation, Shenzhen, China. }
\thanks{$^1$Kaiwen Xue and Huihuan Qian are with the Shenzhen Institute of Artificial Intelligence and Robotics for Society (AIRS), The Chinese University of Hong Kong, Shenzhen, Guangdong, China}
\thanks{$^2$Tong Shen, Tianqi Zhang, Kai Yuan, Kaiwen Xue and Huihuan Qian are also with the School of Science and Engineering, The Chinese University of Hong Kong, Shenzhen,  Guangdong, China.}
\thanks{$\star$ Tong Shen, Tianqi Zhang, and Kai Yuan contribute equally.}
\thanks{$\dagger$The corresponding author is Huihuan Qian.}
}

\begin{document}
\maketitle
\thispagestyle{empty}
\pagestyle{empty}

%%%%%%%%%%%%%%%%%%%%%%%%%%%%%%%%%%%%%%%%%%%%%%%%%%%%%%%%%%%%%%%%%%%%%%%%%%%%%%%%
\begin{abstract}
Implementing fully automatic unmanned surface vehicles (USVs) monitoring water quality is challenging since effectivelt collecting environment data while keeping the platform stable and environmental-friendly is hard to approach.
To address this problem, we construct a USV that can automatically navigate an efficient path to sample water quality parameters in order to monitor the aquatic environment. 
The detection device needs to be stable enough to resist a hostile environment or climates while enormous volumes will disturb the aquaculture environment. Meanwhile, planning an efficient path for information collecting needs to deal with the contradiction between the restriction of energy and the amount of information in the coverage region.
To tackle with mentioned challenges, we provide a USV platform that can perfectly balance mobility, stability, and portability attributed to its special round-shape structure and redundancy motion design. For informative planning, we combined the TSP and CPP algorithms to construct 
an optimistic plan for collecting more data within a certain range and limited energy restrictions.
We designed a fish existance prediction scenario to verify the novel system in both simulation experiments and field experiments.
The novel aquaculture environment monitoring system significantly reduces the burden of manual operation in the fishery inspection field. Additionally, the simplicity of the sensor setup and the minimal cost of the platform enables its other possible applications in aquatic exploration and commercial utilization.
\end{abstract}

%%%%%%%%%%%%%%%%%%%%%%%%%%%%%%%%%%%%%%%%%%%%%%%%%%%%%%%%%%%%%%%%%%%%%%%%%%%%%%%%
\section{Introduction}
Aquaculture has gradually attached more importance since fish demand has sharply increased. China is one of the most crucial aquaculture countries in the world, but there are still some problems with low digitalization, intelligence, and labor productivity \cite{c1}. Detecting the aquatic environment and variation of fish species is the key to utilizing and protecting aquatic resources \cite{c2}. For example, poor water quality can result in low profit, low product quality, and potential human health risks \cite{c3}. For scientific research, water quality parameters can help scientists predict the distribution of aquatic animals \cite{c4}. The applications cover traditional fishing and aquatic agriculture as well as environmental protection and aquatic research. Smart aquaculture is considered the third green revolution. It is a deep integration of modern information technology to achieve intelligent production and intelligent decision for improving the quality and productivity of aquaculture \cite{c1}. Even though various advanced modern technology has been widely applied to the aquaculture industry, the aquaculture equipment market still have room for improvement.
\begin{figure}[htp]
	\centering
	\includegraphics[width=\linewidth]{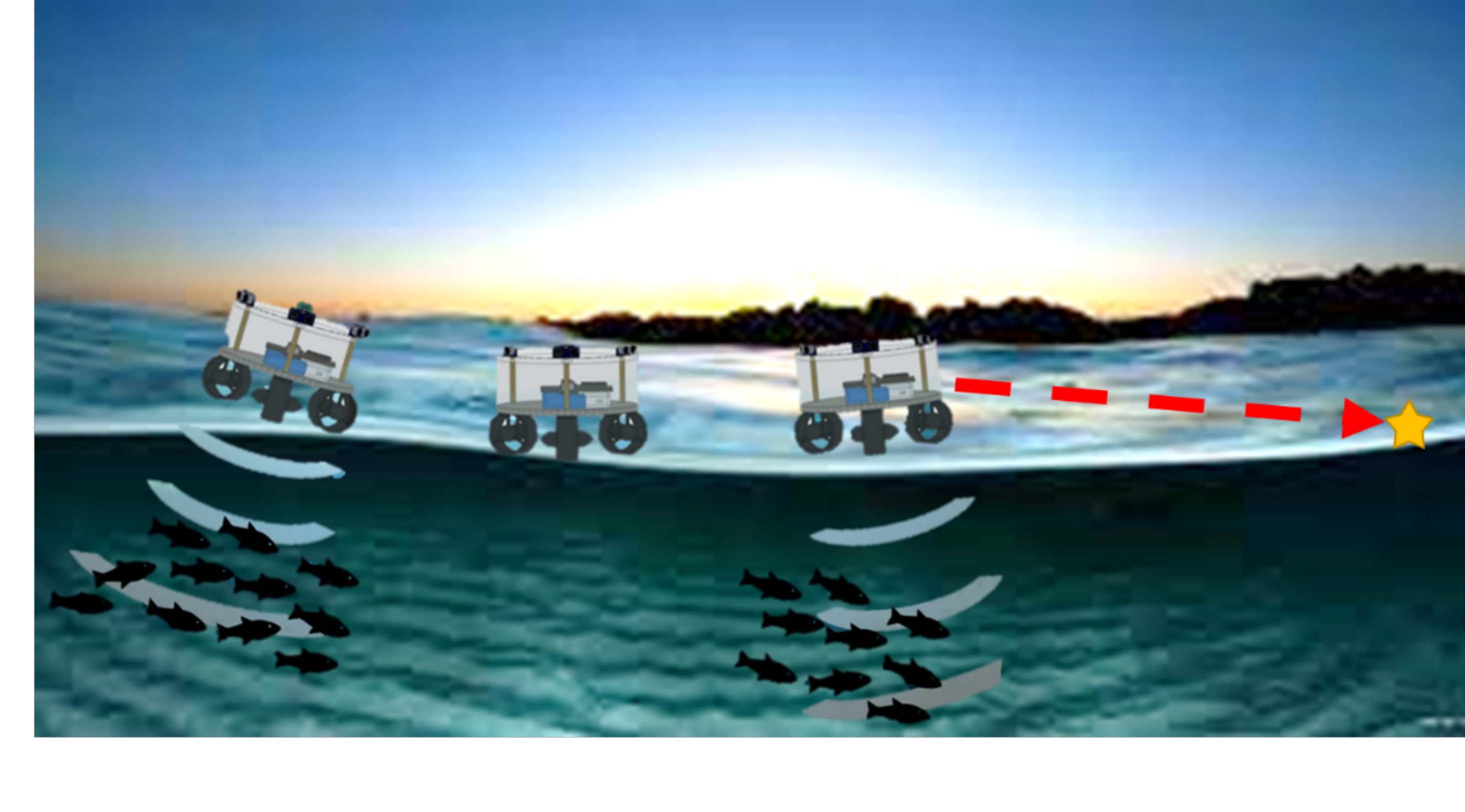}
	\caption{Conceptual application for USVs monitoring in an open body for fishery.}
	\label{intro-pics}
\end{figure}

In order to satisfy the rapid development of aquaculture farming and research, the water quality monitoring field has already seen advanced technology applications like IoT and robotics USV. To reduce waste and prevent water contamination from excessive feed, IoT solutions have been implemented to achieve automated feeding on fish farms. \cite{c5}. For USV design, HAUCS is capable of collaborative monitoring and decision-making on farms of various scales \cite{c5}. Sea farm inspector adopts the function of inspecting and monitoring aquaculture installations at sea. Nevertheless, some concerns emerge: these platforms have difficulty balancing mobility, stability, and portability. Moreover, the sizable volume and mechanical noise in the aquaculture environment pose a threat to the health of fish \cite{c6}. Even though vital progress has been created, several challenges in aquaculture remain. The professionals in the fish farming sector have previously acknowledged the demand for aquaculture-friendly equipment, taking automaticity, low noise, high efficiency, environmentally friendly usability, and no extra emphasis placed on fish into account \cite{c6}. To fill the knowledge gap mentioned, our study designed a  platform focusing on stability ascribe to its special round design that is diminutive to minimize the disturbance to the aquaculture environment. Based on high-portability, our project provides a special redundancy structure to prevent incidents such as a steering gear breaking down or getting tangled up in gutters or garbage which contributes to the robustness of the USV.

To enhance the efficiency and convenience of collecting water quality data, a novel planning method is inevitable, considering automaticity. Even though many path planning methods have been
applied and verified in water quality monitoring problems,
most of the existing studies consider only a single region
to be examined. Based on water monitoring platforms, an
algorithm called TSP-CPP which determines the optimal path
to fully cover all regions exactly once considering sufficient
power of the platform has been raised to solve multi-regions
path planning problems \cite{path}. However, the study was still at
theoretical and simulation levels without entity application. To fill the knowledge gap, we aim to simplify the algorithm and apply the function on the Oboat platform in a fish prediction application. We plan to design a 2-stage path planning model that implements a coarse-fine function in order to improve the efficiency of exploring the whole water surface. 
Our "Oboat" project contributes to combining TSP-CPP planning into the aquatic monitoring field. For practical application, after inputting several specific coordinates to the computer, Oboat will first roughly go through the points to collect water quality parameters and generate an optimal path according to the water parameters collected and the application scenario that can iterate potential regions within a particular power limitation. Afterward, Oboat will follow the path automatically and sample the water quality parameters along the way.

This project is designed to create a low-cost water quality monitoring platform which is a highly automated USV. The renderings of the project are shown in Figure \ref{intro-pics}. These findings contribute in several ways to prior studies. First, the platform perfectly balances the mobility, stability, and portability of the USV. Second, we provide a novel method to achieve coverage path planning to collect the maximum amount of relevant data within limited energy. Third, we have verified the system through the application of fish detection. According to the environment model, a statistical relationship exists between observations of animal presence or abundance and various attributes of the environment parameters\cite{c7}. Combining the water quality data and the logistic model, we are capable of predicting the presence of a specific species. Simultaneously, the prediction result can be verified by the fish detection radar equipped on Oboat. The present study provides the first comprehensive version of automatically self-planning USV. This project is designed to create a low-cost water quality monitoring platform which is a highly automated USV. The renderings of the project are shown in Figure \ref{intro-pics}. These findings contribute in several ways to prior studies. First, the platform perfectly balances the mobility, stability, and portability of the USV. Second, we provide a novel method to achieve coverage path planning to collect the maximum amount of relevant data within limited energy. Third, we have verified the effectiveness of the information collection system through the application of fish existence prediction. According to the environment model, a statistical relationship exists between observations of animal presence or abundance and various attributes of the environment parameters\cite{c7}. Combining the water quality data and the logistic model, we are capable of predicting the presence of the fish species. Simultaneously, the prediction result can be verified by the fish detection radar equipped on Oboat. The present study provides the first comprehensive version of automatically self-planning USV. 

This paper is structured as follows. Section II thoroughly overviews the system design including hardware and software. Section III argues the informative path planning methods. The fish prediction model used to verify the whole system is explained in Section IV. Experiments towards navigation, sampling, and fish prediction are presented in Section V. Section VI concludes the paper.

\section{System Overview}
\begin{figure}[h!]
	\centering
	\includegraphics[width=\linewidth]{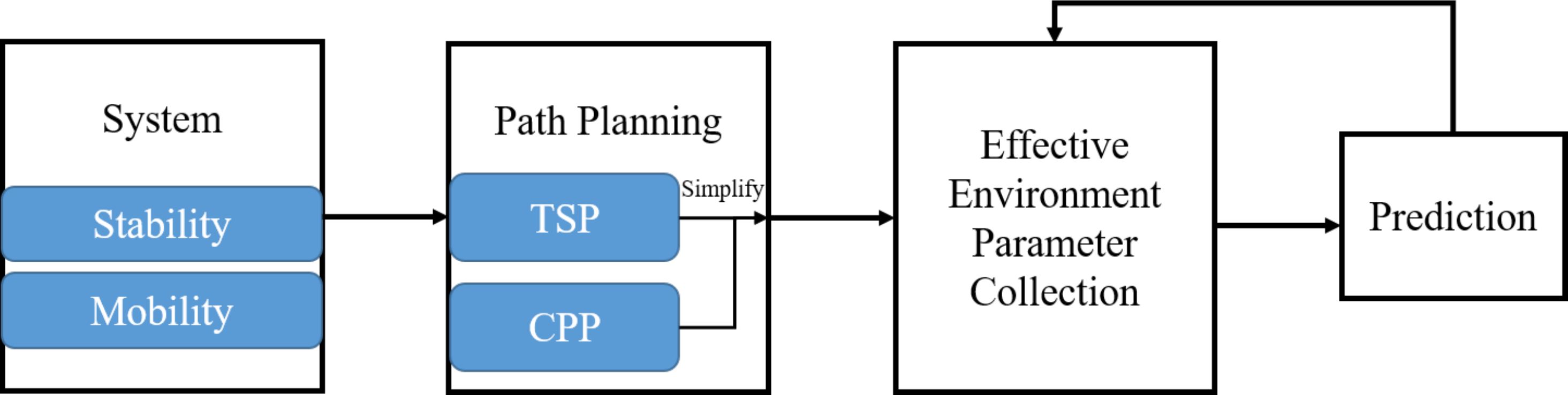}
	\caption{System overview of the proposed path planning and validation schema.}
	\label{fig-system-overview}
\end{figure}
This section presents an overview of our proposed path planning and validation method via fish prediction. The introduction to the platform of the whole method focuses on stability and mobility. The path planning method combines the TSP and CPP methods and simplifies them. The collected environmental parameters are used for fish prediction to verify the effectiveness of the path planning method.

\subsection{Subsystem}

This part introduces the components that make up the entire system, divided into the analysis of the mobile platform and the platform motion algorithm.
\subsubsection{Stability Analysis}
\begin{figure}[h!]
	\centering
	\includegraphics[width=\linewidth]{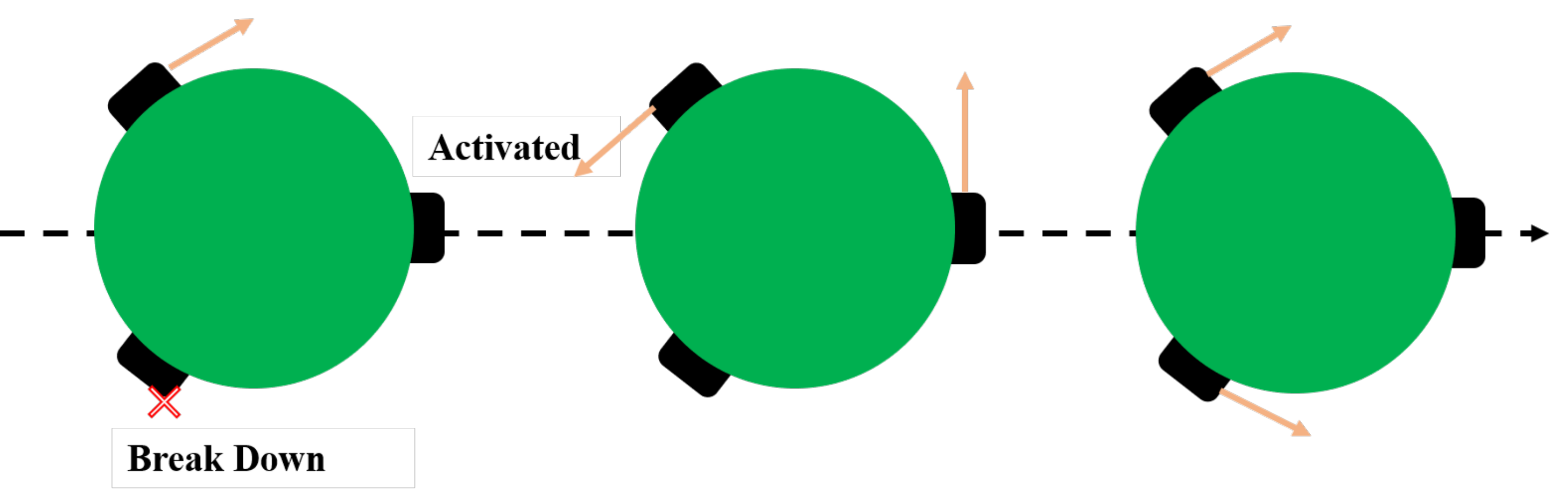}
	\caption{Illustration of platform stability: when one thruster fails during motion, the redundant thrusters are activated and the movement function is restored.}
	\label{fig-yk-stability}
\end{figure}
The platform has good stability and can maintain the motion function in the case of unexpected failure. As shown in Fig. \ref{fig-yk-stability}, when two thrusters work together to complete forward and turn, if one of the thrusters fails, the third redundant thruster is activated at the same time and the working thruster exerts the same thrust. The platform is rotated so that the new thruster pair is oriented in the same direction as the original thruster pair. At this time, the new thruster pair works together to continue the navigation task.

\subsubsection{Mobility Analysis}
In order to allow the motion platform to move according to the planned path, we designed a motion method based on the line of sight algorithm. The target orientation is calculated according to the target position and the current position. By inputting the PID controller together with the current orientation, the thruster of the platform adjusts the thrust to complete the angle control. For details, please refer to ROBIO-A Predictive Method for Site Selection in Aquaculture with a Robotic Platform.

\subsection{Path Planning}
To effectively cover the target water surface to sample the water parameters, we designed a novel coarse-to-fine path planning model which first roughly go through the water surface and then precisely travels the region with greater correlation with the objective scenario. The planning method that combine several regions is a simplified version of the TSP+CPP algorithm that can be applied to the Oboat platform.

\section{Path Planning Based on Coarse-Fine Strategy}
In order to efficiently obtain as much relevant information as possible about the objective application scenario (here we apply the method to the prediction of fish distribution in the waters), we use a coarse-fine approach for path planning. 
\subsection{First Stage -- General Iteration Routes}
First, the zig-zag route is used to randomly obtain the environmental parameter information of the water, and then we pick a threshold and select the sites with factors of importance which we called Region of Interest (ROI). As shown in Fig.4, the red circle represents the first stage start point while the green square represents the end point of \textit{zigzag}. During the path, the environment sensors equipped on the USV platform \textit{Oboat} collects the water parameters.
\begin{equation}
	y^i=f^i(x)
\end{equation} 
according to the obtained information, where $y^i$ represents the value of the environmental parameter $i$ at position $x$.

The ROI is determined by the factors that matter most in each application. In this project, we apply this planning method to a machine learning application of fish occurrence prediction. The black stars represent the selected sited where the fish occurrences detected by the sonar radar equipped on \textit{Oboat} are greater than the threshold. Then, we define the circumcircle of each selected points cluster as the ROI in this situation, shown as the three circles in different colors.
A continuous distribution of the probability of fish occurrence in waters can be obtained as:
\begin{equation}
	n_{ocr}\sim F(x)
\end{equation}
By obtaining the occurrence distribution $n_{ocr}$, some areas with a high probability of fish occurrence can be obtained.
\begin{equation}
	X_i=\{x\in \chi :F(x)>\epsilon\}
\end{equation}
Where $\epsilon$ is the threshold of interest, $\chi$ is the working space.
\begin{figure}[htp]
	\centering
	\includegraphics[width=\linewidth]{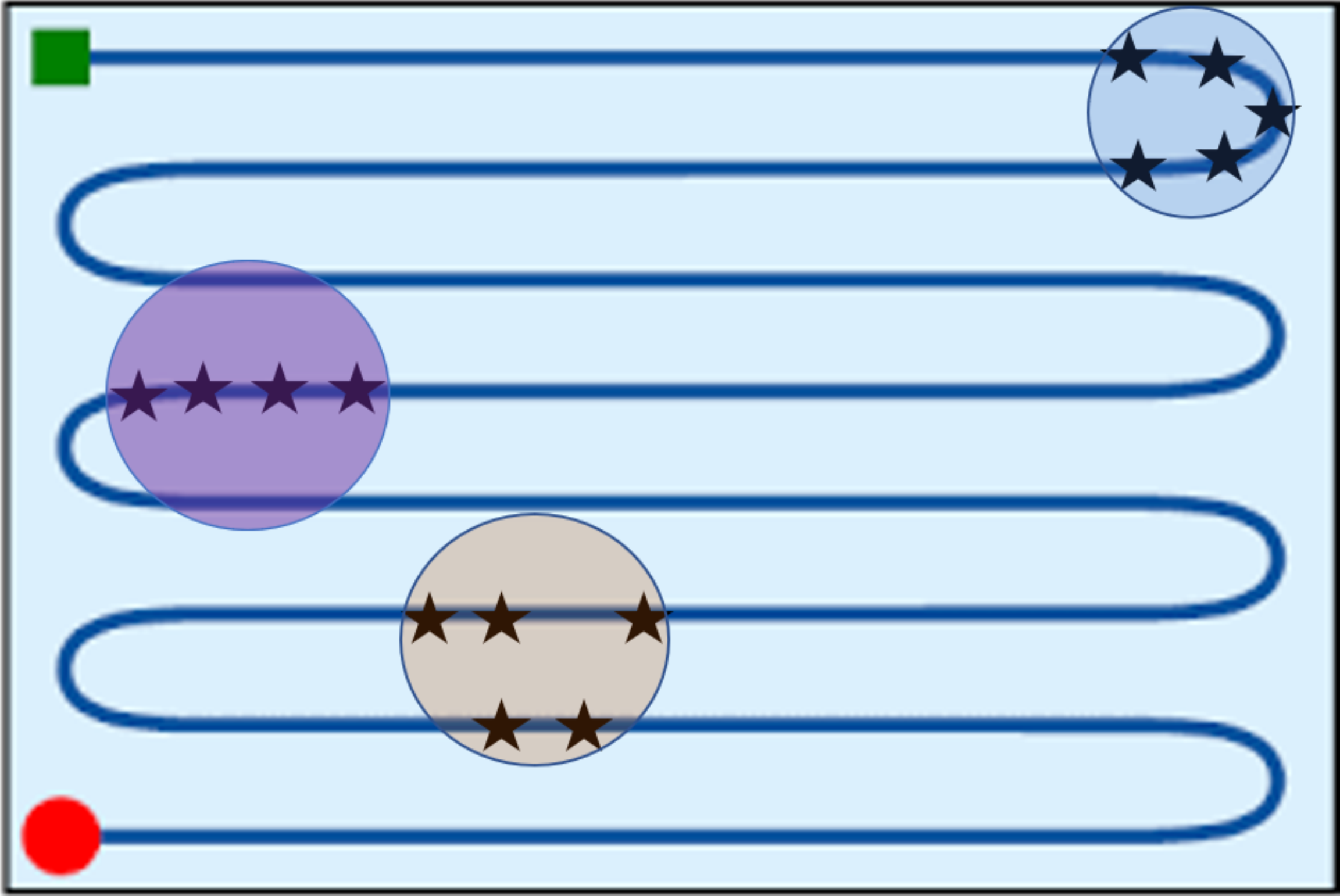}
	\caption{Conceptual zigzag path style of the first stage of path planning.}
	\label{stage1}
\end{figure}

\subsection{K-means Clustering for Site Selection}
\begin{figure}[h]
	\centering
	\includegraphics[width=\linewidth]{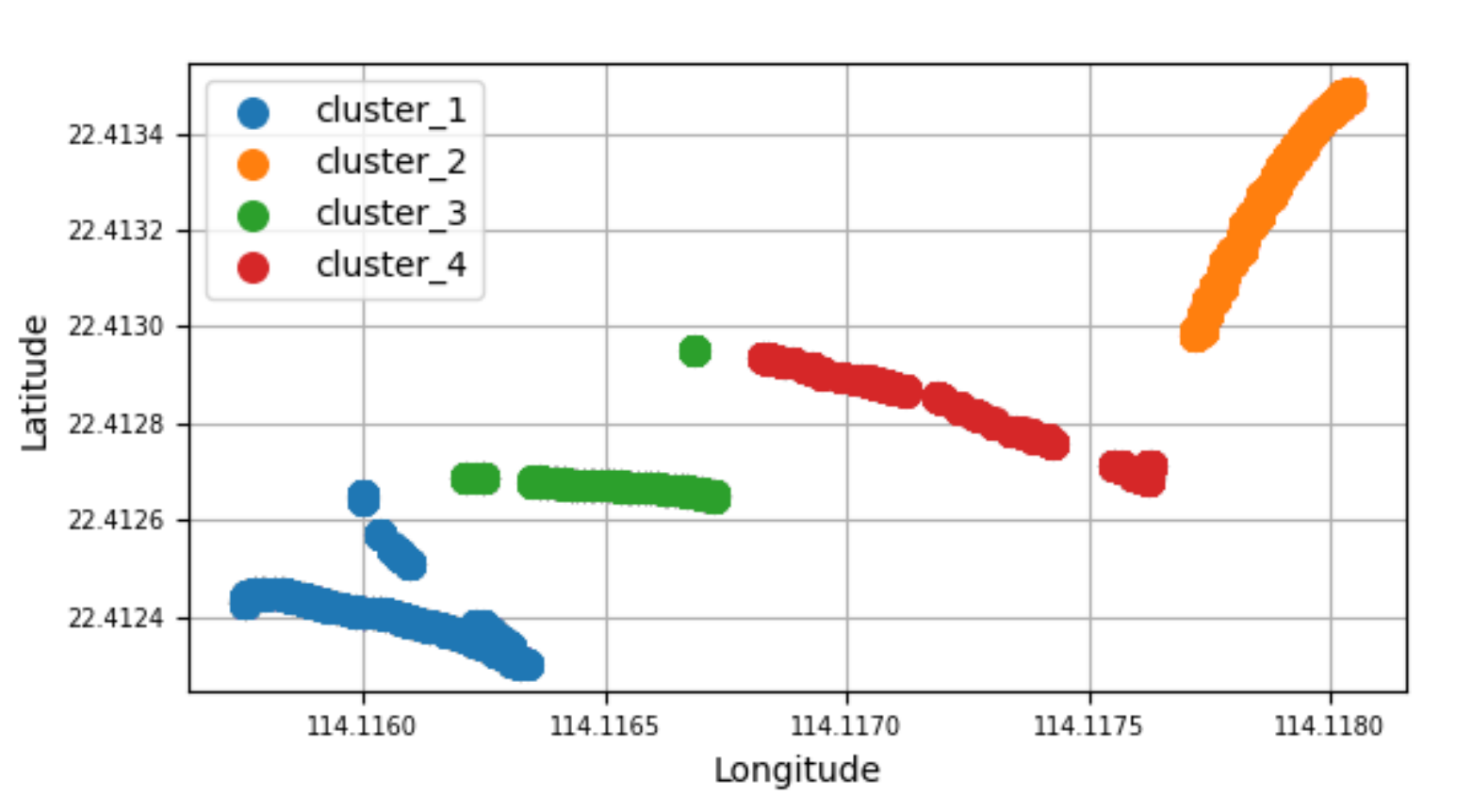}
	\caption{Clustering simulation result.}
	\label{cluster}
\end{figure}

With given probabilities returned by the logistic model, set the threshold to pick out locations with high fish occurrence probability. Record the GPS coordinates of the chosen sites given by the navigation system and form a sample $\mathbf{X}$ with sample size $n$. Each $x_{i}$ in $\mathbf{X}$ represents coordinate. The k-means algorithm divides $\mathbf{X}$ into $k$ disjoint clusters $\mathbf{C}$, each described by the mean, or called the cluster centroid, $c_{j}$ of the samples in the cluster. The K-means algorithm aims to choose centroids that minimize the inertia: 
% $$\sum_{i=0}^{n} \min_{c_{j}\in C } \left ( \left \|x_{j} -c_{j} \right \|  \right )$$
\begin{equation}
	\sum_{i=0}^{n} \min_{c_{j}\in C } \left ( \left \|x_{j} -c_{j} \right \|  \right )
\end{equation}
After the process of clustering, the qualified locations with high probabilities will be divided into several regions. These regions are suitable aquaculture sites that we want for further patrolling and monitoring. A simulation for the clustering result is shown in Fig. \ref{cluster}.

\subsection{Second Stage -- Region Path Planning}
After picking out several ROIs, we need to plan the path between different regions and inside each region.
\subsubsection{Iterate Between Regions}
First, consider each region as a point at their center of circles and we formulate a TSP optimization problem to plan the path for further exploration. The centers of all the points of interest are collected as the vertices $V$ of a graph, and each of the two vertices is connected, creating a fully connected graph $G$. Then the TSP problem is:
\begin{align}
	\label{TSP} &\min \sum_{i}\sum_{j}c_{ij}y_{ij}\\
	\nonumber s.t &\sum_{j}y_{ij}=1, j=0,1,...,n-1\\
	\nonumber	&\sum_{i}y_{ij}=1, i=0,1,...,n-1
\end{align}
where $c_{ij}$ is the distance between $v_i$ and $v_j$, $y_{ij}$ is a decision variable as follow:

\begin{equation}
	y_{ij}=
	\left\{
	\begin{array}{lr}
		1, \text{$v_j$ is visited after $v_i$}, &  \\
		0, \text{otherwise} &  
	\end{array}
	\right.
\end{equation}
Solving problem \ref{TSP} returns the planned path $P$ for further exploration.

\subsubsection{Intra-regional planning}
Inside each region, we plan a zigzag iteration path to further explore ROI. By equally dividing the circumcircle into 8 pieces on the line, we construct the path according to the single region in Fig.6(a).

% \begin{figure}[htp]
% 	\centering
% 	\includegraphics[width=\linewidth]{single_stage2.PNG}
% 	\caption{A demonstration of iteration in a single region.}
% 	\label{single_stage2}
% \end{figure}

Combining the two steps, we can construct the completed path of stage 2 shown in Fig.6(b), with the red star represent the starting point.
\begin{figure}[] 
	\centering  
	\subfigure[A simulation of iteration in a single region.]{
		\label{single_stage2}
		\includegraphics[width=0.9\linewidth]{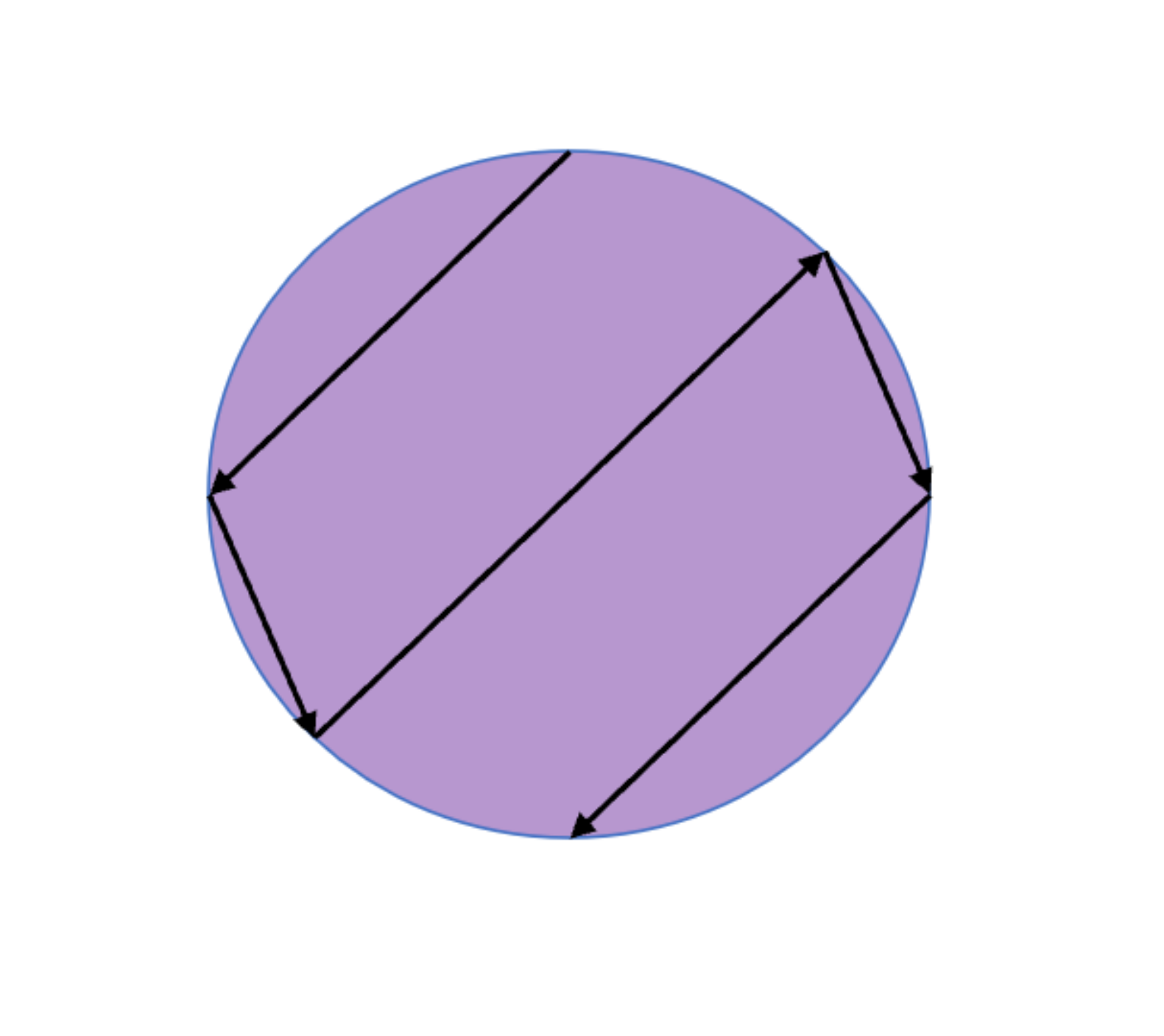}}
	\hspace{-10mm}
	\subfigure[Clustering results and circle bordering]{
		\label{single_stage3}
		\includegraphics[width=0.9\linewidth]{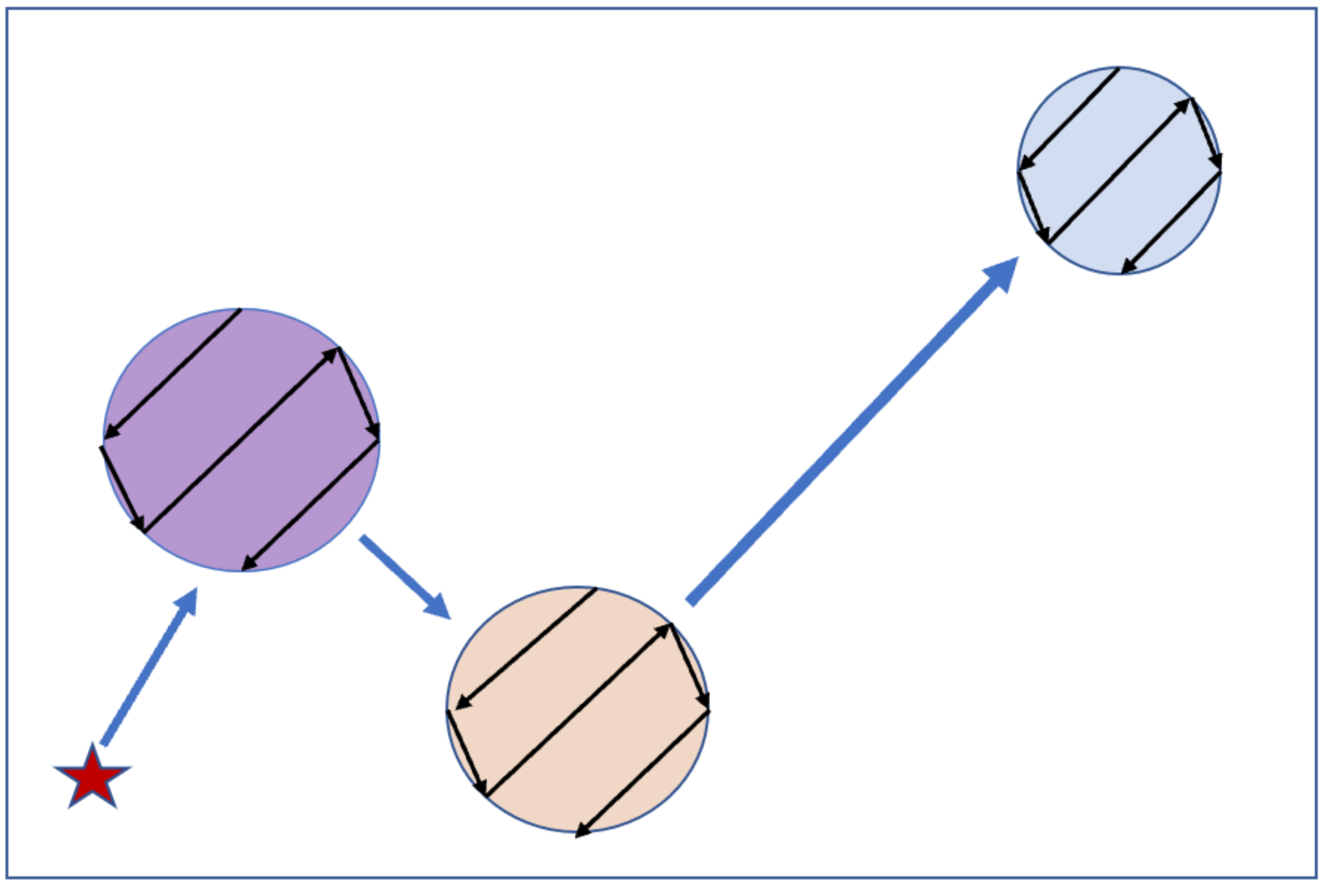}}
	\caption{The completed path of stage 2 path planning.}
	\label{general_path}
\end{figure}
% \begin{figure}[htp]
% 	\centering
% 	\includegraphics[width=\linewidth]{stage2.png}
% 	\caption{The completed path of stage 2 path planning.}
% 	\label{single_stage2}
% \end{figure}
\section{Environment Information Monitoring}
% \begin{itemize}
% \item introduce machine learning methods, reference to scikit-learn package
% \item parameters chosen, PH, DO, reference and tables
% \item Benchmark from radar fish detection
% \end{itemize}
Coarse-fine path planning contributes to more effectively fish habitat monitoring. At aquaculture site, water parameters are key factors that effect fish amount. This part describes the process to fetch water parameters  and to receive fish occurrence information. To validate the collected data, we use water parameters as input data and fish occurrence information as benchmark to train a machine learning model. The model is then utilized for fish occurrence probability prediction, which can reflect whether the environment conditions are suitable for fish reproduction in a certain region.
\subsection{Data Collection}
\subsubsection{Water Parameters' Fetching}
Various water parameters related to fish habitat were studied \cite{Water_parameter}, and PH, temperature, and TDS (turbidity) are three most important among them that control fish amount and growth \cite{c9}. To sample them in real time without deviations, corresponding environment sensors are embedded onboard, uploading data to the mircocomputer. These data will be used as input vectors to train machine the learning model.
\subsubsection{Fish Occurrence Detection}
\begin{figure}[h]
	\centering
	\includegraphics[width=0.9\linewidth]{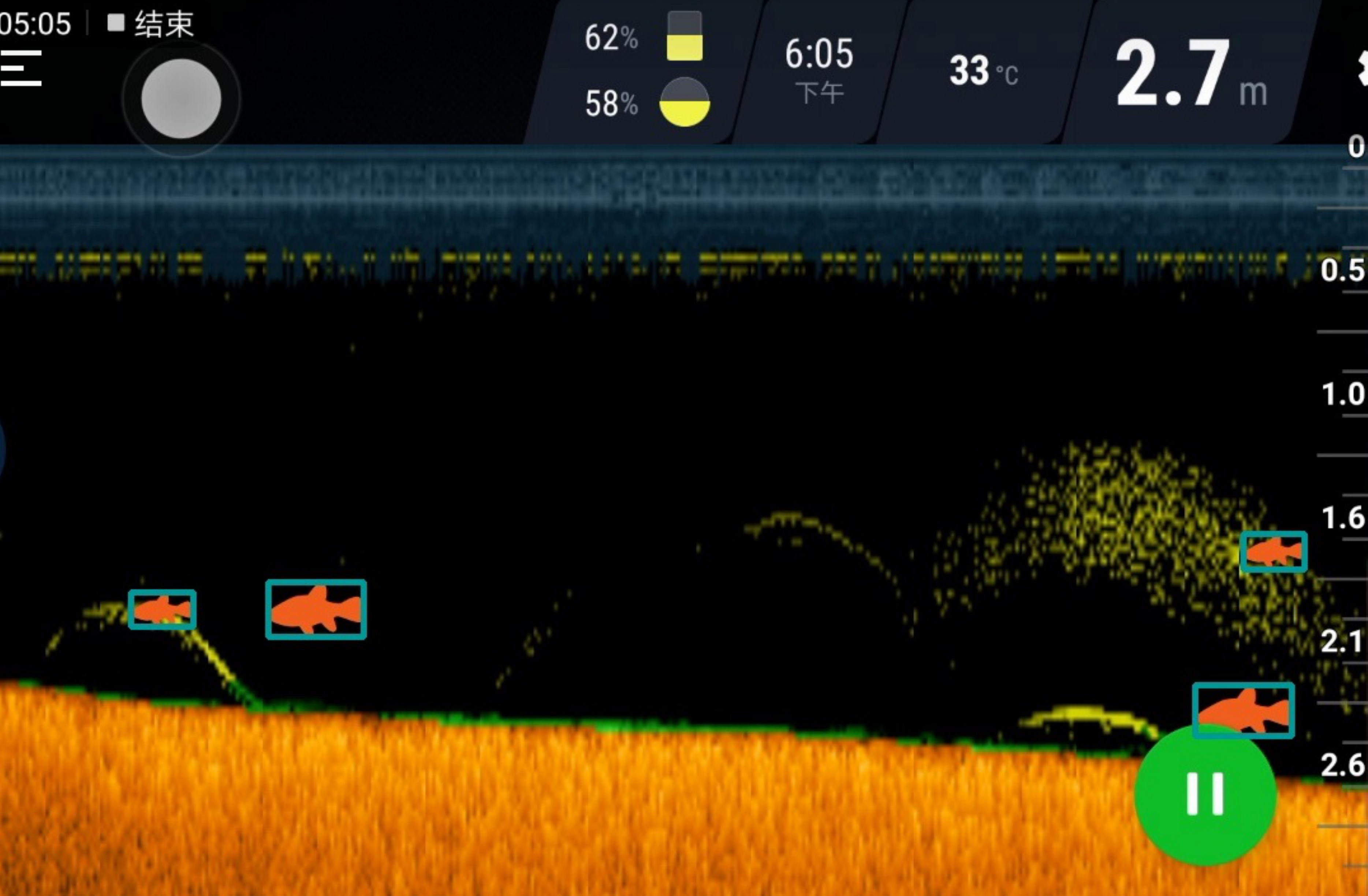}
	\caption{Fish detection radar output.}
	\label{exp-fish}
\end{figure}

 We use fish detection radar to obtain ground truth value $y$ (0 or 1). Fish detection radar determines the water depth, underwater terrain, and fish present through ultrasonic signal transmission. Then it will transfer reflected signals into a 2D dynamic image. The image is updated from right to left (rolling from right to left) on the screen at every millisecond. Fish occurrence, size, and other underwater information are demonstrated in Figure \ref{exp-fish}.

\subsection{Data Validation}
\subsubsection{Benchmark Setting}
\begin{figure}[]
	\centering
	\includegraphics[width=0.9\linewidth]{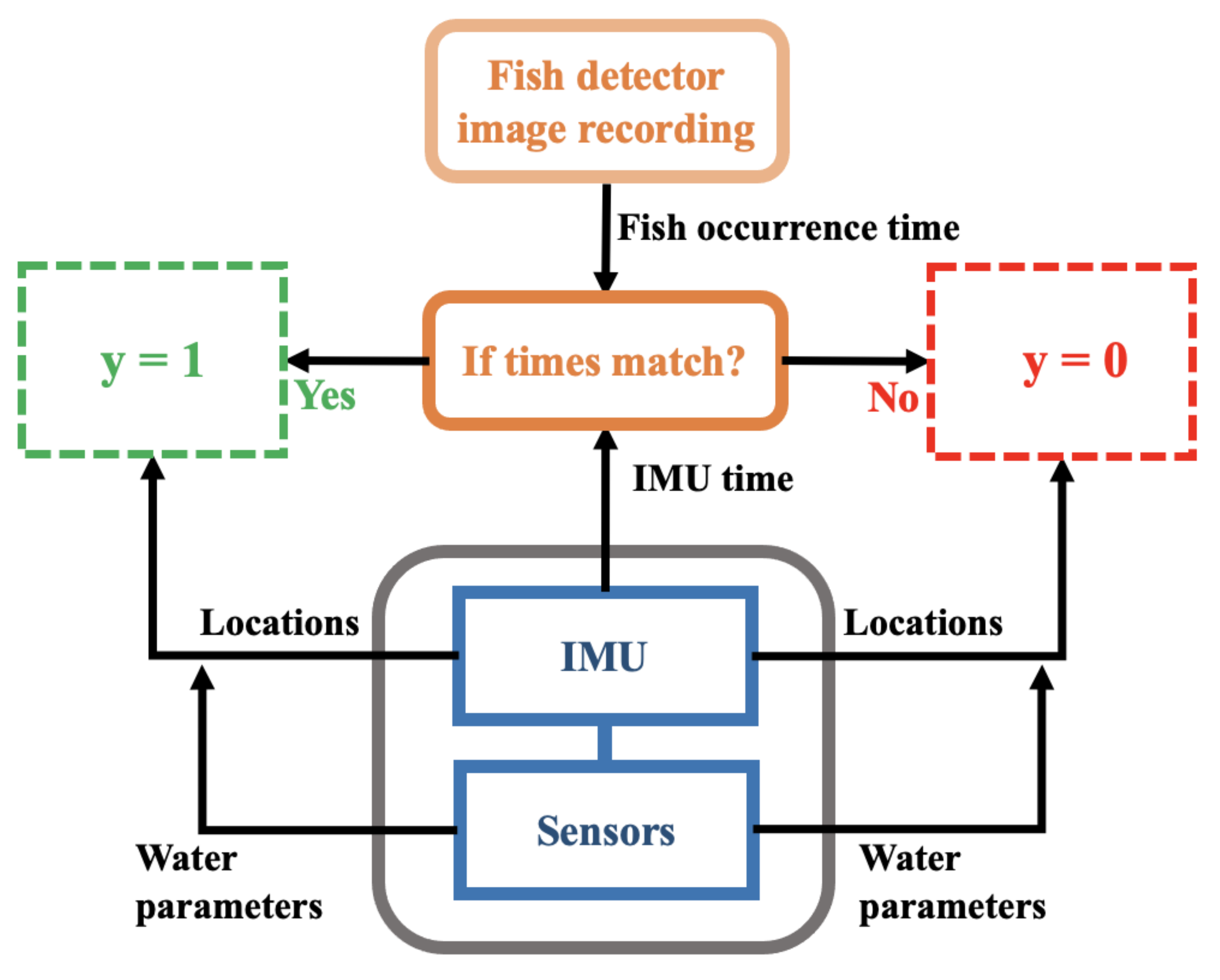}
	\caption{The process of benchmark setting.}
	\label{exp-mark}
\end{figure}
 We have obtained water parameters and fish occurrence inoformation. The next step is to relate ground truth values with input data to fit the model. We use the timestamp as a standard for locating labels. Since the image returned by detection radar is updated from right to left, we first mark down the time when the fish pattern appears at the right border of the screen. The IMU on Oboat will return time and coordinates. If the time of the fish pattern present and the time of IMU are matched, then we set the label of the corresponding location as 1 ($y=1$). Other locations that meet no fish occurrence will be given with label 0. The water parameters receivers are written in the same loop with IMU and output together, therefore the input data can have relationships with ground truth value. The above process is described in the flow chart Figure \ref{exp-mark}.
\subsubsection{Machine learning Model}
Compared with tree-based methods or genetic algorithms, Regression models have generally performed better in predicting species presence \cite{models}. Regression-based models can provide predictions of the probability of occurrence, and the logistic model is frequently applied \cite{c4}. By fitting the sigmoid function, logistic regression can predict the probability of class labels.
Mathematically, the sigmoid function is represented as
\begin{equation}
\label{sigmoid}
\begin{aligned}
	P(y_{i} = 1|X_{i})&=expit(X_{i}w+w_{0})\\
	&=\frac{1}{1+exp(-X_{i}w-w_{0})}\\
	&=\hat{p}(X_{i})
\end{aligned}
\end{equation}
Where $P(y_{i}=1|X_{i})$ is the probability of fish occurrence, $y_{i}$ is the ground truth value, $X_{i}$ is a vector of water quality parameters, $w$ are unknown weights to be estimated. The independent variable is user defined, we can choose the parameters according to the specific problem.

The process of fitting binary logistic regression can be transformed into an optimization problem. The optimization process is related to l2 regularization, shown in the equation (\ref{reg}). Solve the optimization problem by minimizing the cost function in equation (\ref{costf}):
\begin{equation}
\label{reg}
\begin{aligned}
    r(w) = \frac{1}{2}\left \| W \right \| _{F}^{2}=\frac{1}{2} {\textstyle \sum_{i=1}^{n}}  {\textstyle \sum_{j=1}^{K}}  W_{i,j}^{2}
\end{aligned}
\end{equation}

\begin{equation}
\label{costf}
\begin{aligned}
\min_{w}C\sum_{i=1}^{n}(-y_{i}log(\hat{p}(X_{i}))-(1-y_{i})log(1-\hat{p}(X_{i})))+r(w)
\end{aligned}
\end{equation}
% $$\min_{w,c} \frac{1}{2}w^{T}w+C\sum_{i=1}^{n}log[exp(-y_{i}(X_{i}^{T}w+c))+1]$$  

\section{Experiments and Results}
This section presents experiments to verify the system functionality and the path planning algorithm and our prediction method. 
\subsection{Experiment Scenarios}
\begin{figure}[]
	\centering
	\includegraphics[width=0.9\linewidth]{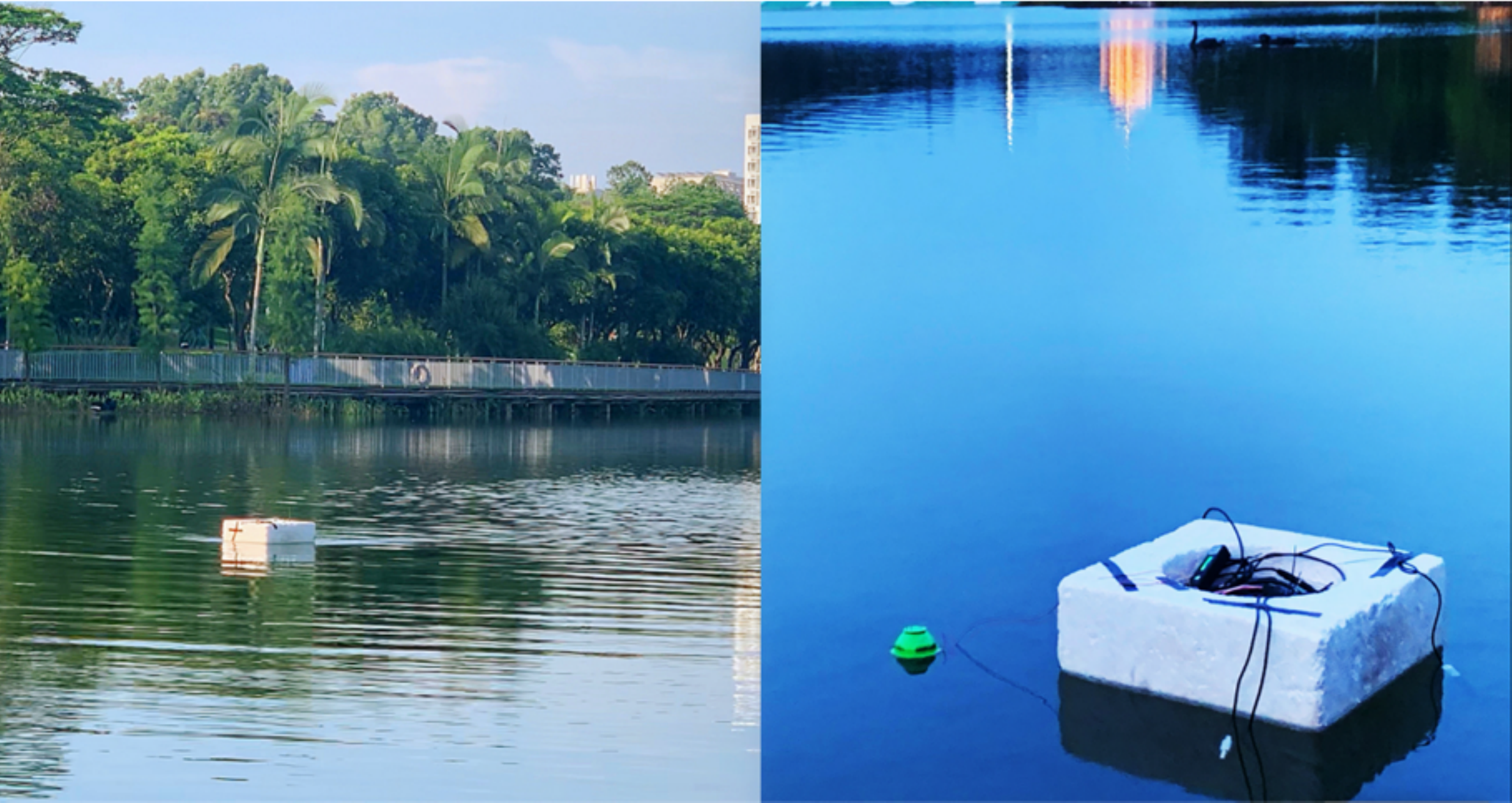}
	\caption{Practical experiment scenarios in an outdoor open water body.}
	\label{lake_view}
\end{figure}

The field experiment was conducted in an outdoor lake,shown in Figure. \ref{lake_view}. Our robot automatically navigated the lake and collected environmental information. The onboard sensors sampled the environmental information of pH, temperature, dissolved oxygen, and turbidity density along the navigation path. Together with the radar fish detection data, the prediction model could be trained and tested for future application. 
 .

\subsection{Path Planning}
In this section, we utilized the environmental parameters obtained from the first stage navigation, generated an efficient path for iterating potential regions, and showed the feasibility of the Oboat platform.
\subsubsection{General Iteration}
% \begin{figure}[h]
% 	\centering
% 	\includegraphics[width=\linewidth]{navigation.png}
% 	\caption{Theoretical cruise track of the navigation path (red); Practical navigation path in experiment}
% 	\label{lake_path}
% \end{figure}

\begin{figure}[] 
	\centering  
	\subfigure[Theoretical cruise track of the navigation path (red); Practical navigation path in experiment]{
		\label{lake_path}
		\includegraphics[width=0.9\linewidth]{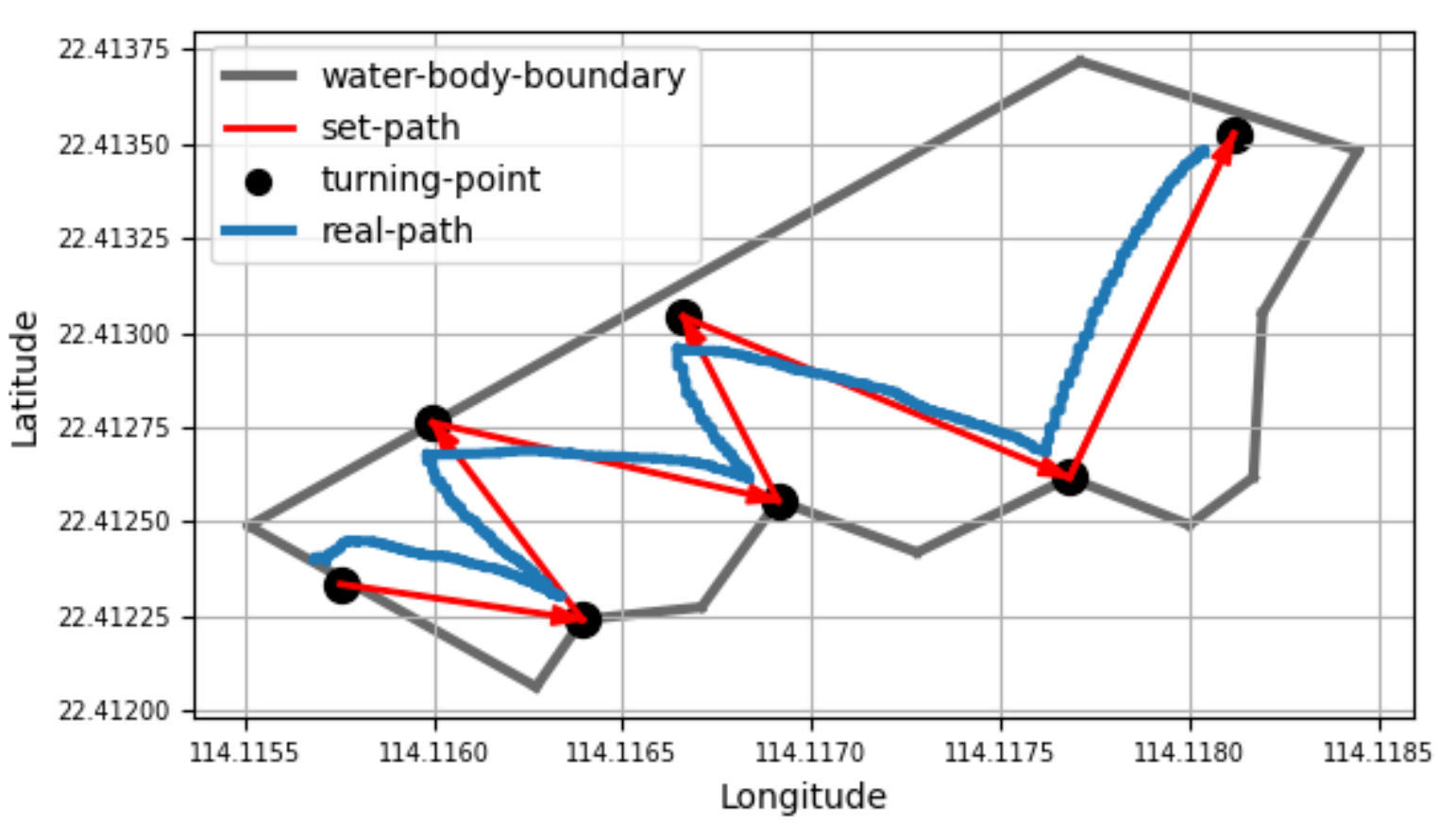}}
	\hspace{-10mm}
	\subfigure[Clustering results and circle bordering]{
		\label{clsuter_site}
		\includegraphics[width=0.9\linewidth]{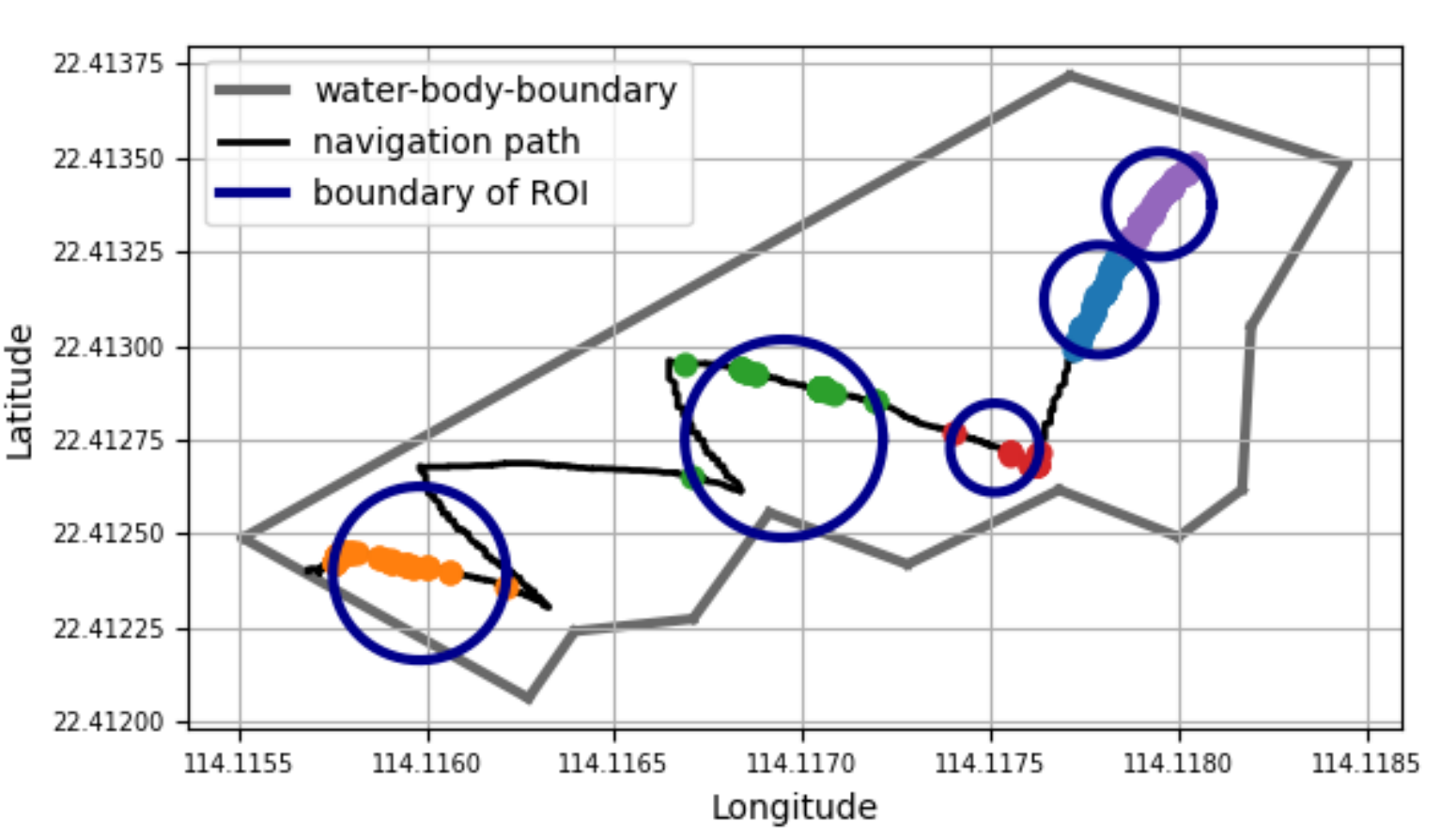}}
	\caption{ROIs found after general iteration}
	\label{general_path}
\end{figure}

We collected seven GPS coordinates on the water area and set them as the turning points for the robot's navigation path. As shown in Fig. \ref{lake_path}, the black dots are turning points, the red arrows are reference sailing routes, and the blue curves are experiment navigation data. From the start point at the bottom left, the robot dynamically tracks the reference paths. It successfully tracks every turning point and navigates to the next turning point until it reaches the endpoint.

\subsubsection{Cluster of Fish Occurrence Sites}
% \begin{figure}[h]
% 	\centering
% 	\includegraphics[width=\linewidth]{circles.png}
% 	\caption{Clustering results and circle bordering.}
% 	\label{lake_path}
% \end{figure}
% \begin{figure}[h] 
% 	\centering  
% 	\subfigure[Clusters of High Frequency Fish Occurrence Sites]{
% 		\label{sub.1}
% 		\includegraphics[width=0.9\linewidth]{cluster_site.png}}
% 	\hspace{-10mm}
% 	\subfigure[Circles bordering ROIs]{
% 		\label{sub.2}
% 		\includegraphics[width=0.9\linewidth]{circles.png}}
% 	\caption{Clustering results.}
% 	\label{clsuter_site}
% \end{figure}
During the first stage of navigation, fish detection radar also recorded fish occurrence information. Fish occurrence sites are marked on the path and their coordinates are points in sample $\mathbf{X}$ which will be used for clustering. In this experiment, we are eager to find 5 ROIs, meaning there are 5 clusters. Each cluster is bordered by the smallest circle that can cover all the fish occurrence points in the interest region. The clustering results are shown in Fig. \ref{clsuter_site}.

% \begin{figure}[b]
% 	\centering
% 	\includegraphics[width=\linewidth]{cluster_site.png}
% 	\caption{Cluster of High Frequency Fish Occurrence Sites}
% 	\label{cluster_site}
% \end{figure}

\subsubsection{Regions of Interests Iteration Path Planning}
After getting ROIs, we developed the second stage path for further detailed iteration through area that have higher possibility of fish occurrence. Fig. \ref{single_stage2} shows the detailed dots and the red arrows represent the order travels between them.
% \begin{figure}[h]
% 	\centering
% 	\includegraphics[width=\linewidth]{zigzag.PNG}
% 	\caption{Second stage path planning between and inside ROIs}
% 	\label{single_stage2}
% \end{figure}

\subsubsection{Intra-Regional Planning Validation}
We choose a ROI in the lake and generated the path inside the region, and we evaluated the feasibility of the path on the Oboat platform by field experiment. The result is shown in Fig. \ref{itra_plan}. The USV generally travel through the designed path considering the natural 10-meter error of the GPS which proved that the novel path planning method is suitable for Oboat platform.
\begin{figure}[] 
	\centering  
	\subfigure[Second stage path planning between and inside ROIs]{
		\label{single_stage3}
		\includegraphics[width=0.9\linewidth]{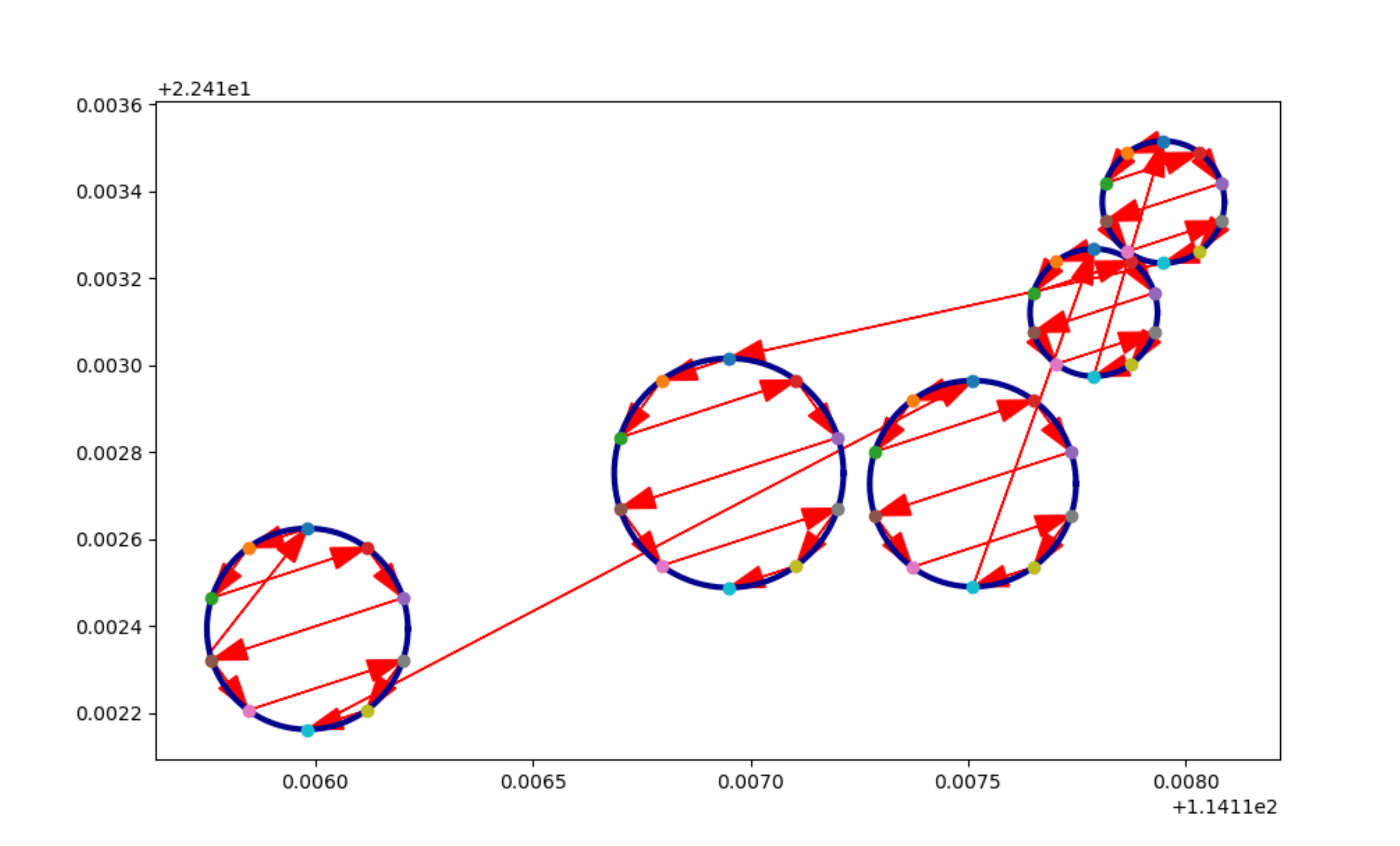}}
	\hspace{-10mm}
	\subfigure[Theoretical and practical path inside one ROI (Black dots: destinations, Red arrows: theoretical path, Blue line: practical path in field experiment)]{
		\label{clsuter_site}
		\includegraphics[width=0.7\linewidth]{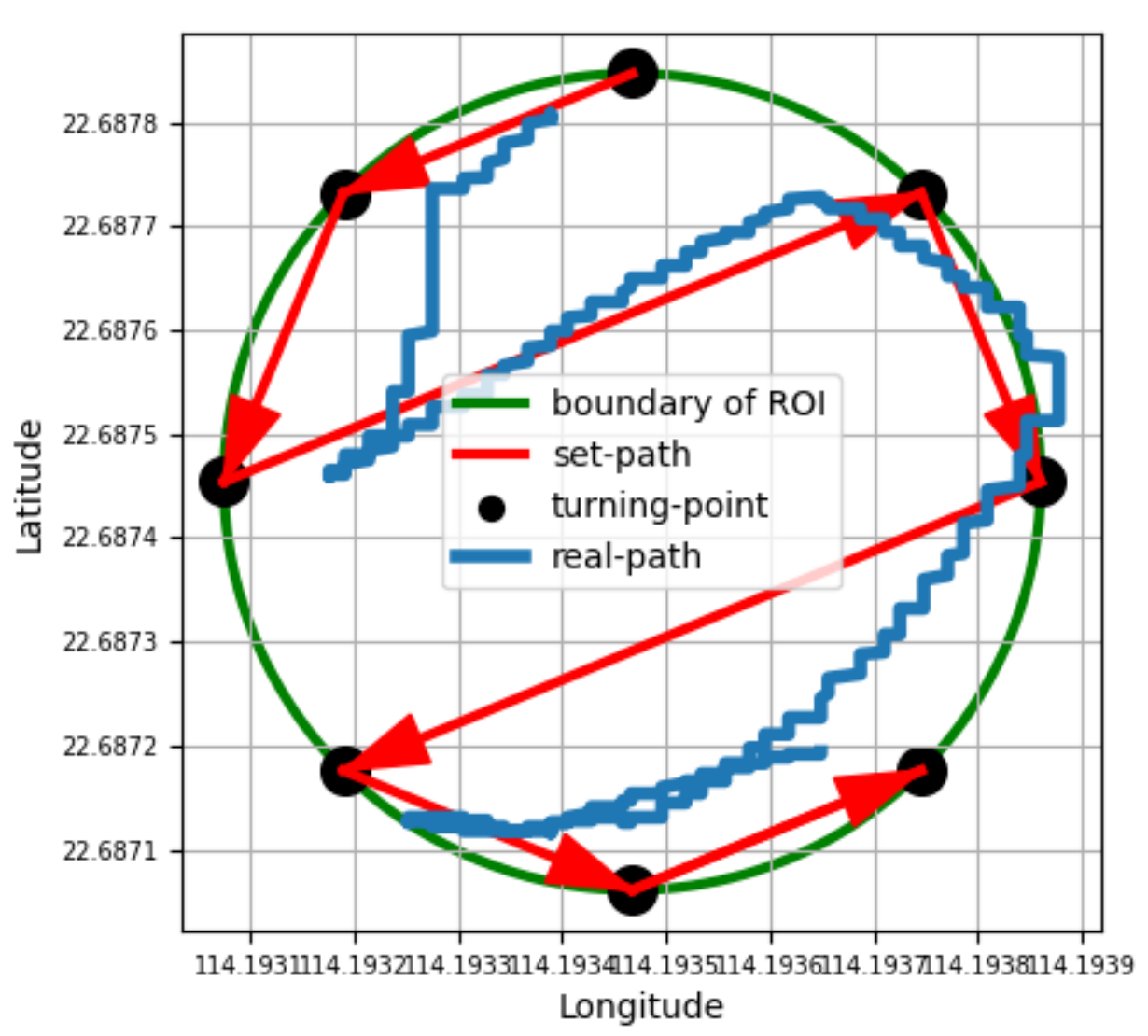}}
	\hspace{-15mm}
	\subfigure[Fish occurrence along the navigation path in the training ROI (left); 3D probability distribution (right).]{
		\label{plan}
		\includegraphics[width=\linewidth]{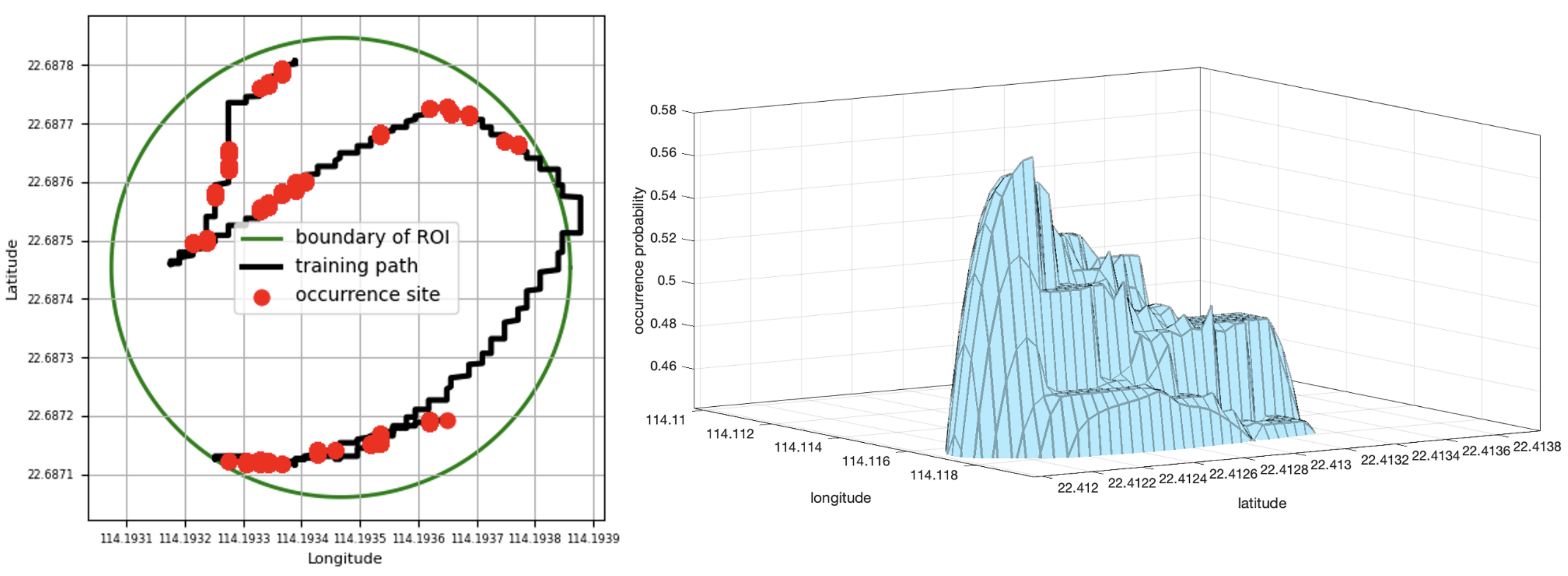}}
	\caption{Demonstration of second stage path planning and future prediction}
	\label{second_path}
\end{figure}

% \begin{figure}[ht]
% 	\centering
% 	\includegraphics[width=0.8\linewidth]{itra_plan.png}
% 	\caption{Theoretical and practical path inside one ROI (Black dots: destinations, Red arrows: theoretical path, Blue line: practical path in field experiment)}
% 	\label{itra_plan}
% \end{figure}
\subsection{Further Plan for Prediction Experiments}
\subsubsection{Training Set Selection and Model Fitting}
We decide to use points in one of the ROIs as the training set. To select the intending ROI, we will integrate K-fold Cross-validation method for training set's evaluation. The selected ROI used as training set will give information about fish occurrence. The ground truth values $y_train = 0$ and $y_train = 1$ (representing fish occurrence) were labeled to each point through the process mentioned in benchmark setting, and fish occurrence sites are shown as red dots along the path in Fig. \ref{plan} (left). Combined with water parameters on the path that are used as input vector $X$, we hope to derive an actual weight vector $w$ for sigmoid function \ref{sigmoid}. The fitted model will be used for further prediction. We will use confusion matrix to calculate precision and accuracy of the model.
\subsubsection{Prediction Results and Validation}
The constructed machine learning model will be used to predict fish occurrence probabilities in the rest ROIs that are not used for training. These probabilities will be used for surface fitting, that is, we build a 3D probability distribution function for each testing ROI. A simulation is presented in \ref{plan} (right).  Fish detection radar provides ground truth value for prediction validation, and same evaluation method mentioned in training process will be used. If the surface is established, we will be able to justify aquaculture condition in the corresponding region.
% \begin{figure}[]
% 	\centering
% 	\includegraphics[width=\linewidth]{future prediction.png}
% 	\caption{Fish occurrence along the navigation path in the training ROI (left); 3D probability distribution (right).}
% 	\label{plan}
% \end{figure}

\section{CONCLUSIONS}
In this investigation, the aim was to design and implement a low-cost automatically system that focus on aquaculture environment monitoring. This system provides a stable USV that is capable to efficiently collect aquatic parameters showing promising performance in a novel planning within restricted energy. This greatly enhance the efficiency and automaticity in aquaculture inspection field. The work presented here provides one of the first investigations into implementing information planning
with USV to inspect the aquatic environment. In addition,
such methods can be further applied to specific environmental explorations, providing more intelligent applications to the aquaculture industry. In future work, we will verify the whole coarse-to-fine planning by the fish prediction accuracy.

\addtolength{\textheight}{-12cm}   % This command serves to balance the column lengths

%%%%%%%%%%%%%%%%%%%%%%%%%%%%%%%%%%%%%%%%%%%%%%%%%%%%%%%%%%%%%%%%%%%%%%%%%%%%%%%%

%%%%%%%%%%%%%%%%%%%%%%%%%%%%%%%%%%%%%%%%%%%%%%%%%%%%%%%%%%%%%%%%%%%%%%%%%%%%%%%%

%%%%%%%%%%%%%%%%%%%%%%%%%%%%%%%%%%%%%%%%%%%%%%%%%%%%%%%%%%%%%%%%%%%%%%%%%%%%%%%%

\end{document}